\documentclass[12pt]{article}
\usepackage{amsmath}
\usepackage{graphicx,psfrag,epsf}
\usepackage{enumerate}
\usepackage[square,numbers,sort&compress]{natbib}
\usepackage{indentfirst}
\usepackage{hyperref}
\hypersetup{
    colorlinks=true,
    linkcolor=blue,
    filecolor=magenta,      
    urlcolor=cyan,
}
\newcommand{\blind}{0}

\begin{document}

\def\spacingset#1{\renewcommand{\baselinestretch}%
{#1}\small\normalsize} \spacingset{1}


\if0\blind
{
  \title{\bf CAD: Memory Efficient Convolutional Adapter for Segment Anything}
  \author{Joohyeok Kim, Joonhyeon Song, Seohwan Yun, Seongho Yoon, Sangmin Lee \\
    School of Information Convergence, Kwangwoon University\\}
  \maketitle
} \fi

\if1\blind
{
  \bigskip
  \bigskip
  \bigskip
  \begin{center}
    {\LARGE\bf Title}
\end{center}
  \medskip
} \fi

\bigskip
\begin{abstract}
The Foundation model for image segmentation, Segment Anything (SAM), has been actively researched in various fields since its proposal. Various researches have been proposed to adapt SAM to specific domains, with one notable approach involving the addition and training of lightweight adapter modules. While adapter-based fine-tuning approaches have reported parameter efficiency and significant performance improvements, they face a often overlooked issue: the excessive consumption of GPU memory relative to the number of trainable parameters. Addressing this issue, this paper proposes a memory-efficient parallel convolutional adapter architecture. This architecture connects in parallel with SAM's image encoder, eliminating the need to store activations and gradients of the image encoder during model training. Our proposed architecture demonstrated competitive experimental results while using less than half the GPU memory compared to SAM Adapter, indicating its value as an alternative to simple decoder fine-tuning when hardware limitations preclude adapter-based learning. Our code implementation is available at \href{https://github.com/Kyyle2114/Convolutional-Adapter-for-Segment-Anything}{our github}.
\end{abstract}

\spacingset{1.45}
\section{Introduction}

In 2023, a foundation model for image segmentation called Segment Anything (SAM) \cite{SAM} was introduced. Befitting its status as a foundation model trained on large-scale segmentation datasets, SAM demonstrates excellent zero-shot performance, leading to its active research across various domains. Although SAM demonstrates impressive zero-shot performance, it still faces limitations as a foundation model. Consequently, model fine-tuning is essential to achieve optimal performance in novel domains.

To efficiently train heavy models, various Parameter-Efficient Fine-Tuning (PEFT) methods have been proposed. One such approach is the adapter-based method, which involves adding new modules to the existing heavy model. This method freezes the weights of the heavy backbone network to prevent them from being trained, while adding lightweight modules within the backbone network. It then focuses on training only the necessary components, such as the internal lightweight modules and classification head. 

Adapter modules possess a small number of parameters, yet training with these additional parameters alone can yield performance comparable to or surpassing that of training whole model \cite{MED-SA}. This method of incorporating adapter module has been applied to various architectures with transformer backbones, such as SegFormer, SETR \cite{EVP} and SAM \cite{MED-SA, SA, RSAMSEG}. These applications have reported performance improvements across diverse domains.

While fine-tuning with adapters is undoubtedly parameter-efficient, there is a significant issue that often goes unemphasized: GPU memory constraints. Currently, most deep learning models are trained on GPUs. To train a deep learning model on a GPU, it is necessary to load various elements onto GPU memory, including batch data, model parameters, activations, and gradients. As the model capacity and complexity increase, GPU memory usage increases correspondingly. Consequently, heavy models like transformers may require the use of very small batch sizes on standard GPUs, or may not be trainable at all.

\begin{table}[hbt!]
\centering
\begin{tabular}{ccc}
\hline
Module           & Trainable Parameters & GPU Memory Usage \\ \hline
SAM Decoder      & 4,058,340            & 14,113 MiB \\ \hline
SAM Adapter      & 4,788,740            & 36,871 MiB \\ \hline
SAM Conv Adapter & 5,831,908            & 17,697 MiB \\ \hline
\end{tabular}
\caption{GPU memory usage for each module (ViT-B, batch size = 4).}
\label{tab:my-table}
\end{table}

Due to the nature of adapters, lightweight modules are inserted within the transformer backbone. To train these adapters using backpropagation, it is necessary to store the transformer's activations and gradients in GPU memory. This scenario demands a disproportionately large amount of GPU memory relative to the number of trainable parameters. Table 1, 2 illustrates the number of trainable parameters and GPU memory usage for each module during model training. SAM adapter trains both the adapter within the transformer and the SAM's mask decoder. SAM Conv Adapter refers to the architecture proposed in this paper.

\begin{table}[hbt!]
\centering
\begin{tabular}{ccc}
\hline
Module           & Trainable Parameters & GPU Memory Usage \\ \hline
SAM Decoder      & 4,058,340            & 14,329 MiB \\ \hline
SAM Adapter      & 6,091,908            & 42,623 MiB \\ \hline
SAM Conv Adapter & 5,831,908            & 15,457 MiB \\ \hline
\end{tabular}
\caption{GPU memory usage for each module (ViT-H, batch size = 2).}
\label{tab:my-table2}
\end{table}

To address this issue, we propose a parallel convolutional adapter architecture for SAM, capable of reducing GPU memory usage. The proposed architecture connects parallelly to the transformer backbone and does not require the gradients or activations of the transformer during training, thus dramatically reducing the GPU memory requirements for training.

Following previous studies that applied adapters \cite{EVP, SA}, we utilized two challenging tasks: Shadow detection and Camouflaged object detection. We trained the proposed architecture alongside SAM Adapter and SAM Decoder on these datasets, evaluating and reporting their respective performances.

\section{Related Works}

\subsection{Segment Anything}

Segment Anything (SAM), proposed in 2023, is a foundation model for image segmentation. The model comprises an image encoder utilizing a Vision Transformer (ViT) \cite{VIT}, a prompt encoder, and a mask decoder, and was trained on its proprietary dataset, SA-1B.

The image encoder processes input images to compute image embeddings. Based on various user-input prompts such as points, boxes, masks, or text, the mask decoder makes the segmentation masks.

The SAM model types can be categorized into three variants based on the vision transformer(image encoder) used: ViT-B(Base), ViT-L(Large), and ViT-H(Huge). The smallest model, ViT-B, has approximately 90 million parameters, while the largest, ViT-H, contains about 600 million parameters.

\subsection{Adapters for PEFT}

As the capacity of deep learning models such as transformers continues to increase, various methods are being proposed to efficiently train these heavy architectures. These methods are commonly referred to as Parameter-Efficient Fine-Tuning (PEFT).

One such approach is the adapter-based method. This method involves adding lightweight modules within the transformer, freezing the transformer's weights, and training only the necessary components such as the adapters and classification head. Notable examples include LoRA (Low Rank Adaptation) \cite{LORA}, proposed for training Large Language Models (LLMs) as shown in Figure 1, and QLoRA \cite{QLORA}, which quantizes the transformer's weights with adding LoRA adapters.

\begin{figure}[h]
    \centerline{\includegraphics[width=8cm]{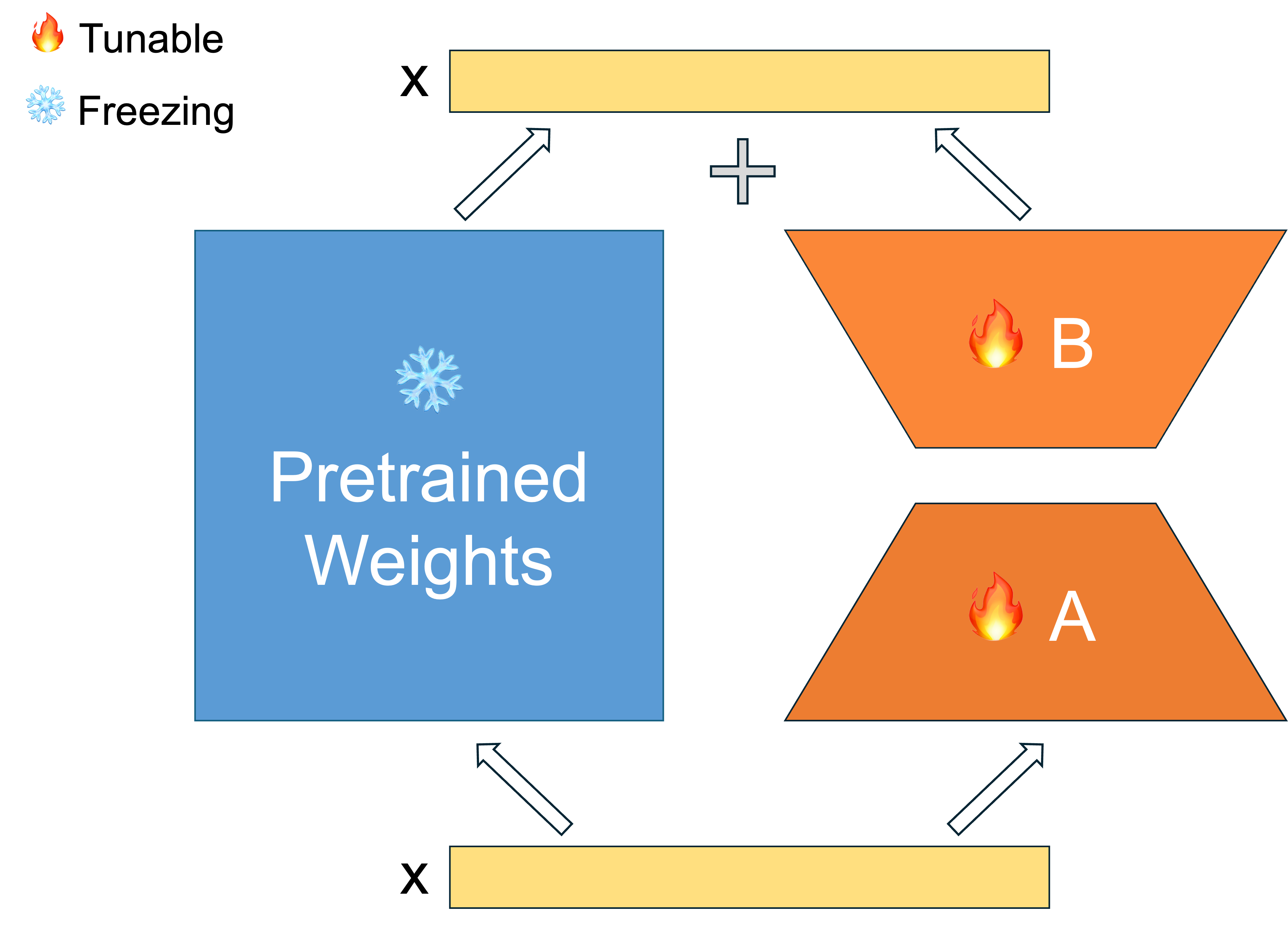}}
    \caption{LoRA architecture.}
    \label{figure_1} 
\end{figure}

\begin{figure}[]
    \centerline{\includegraphics[width=15cm]{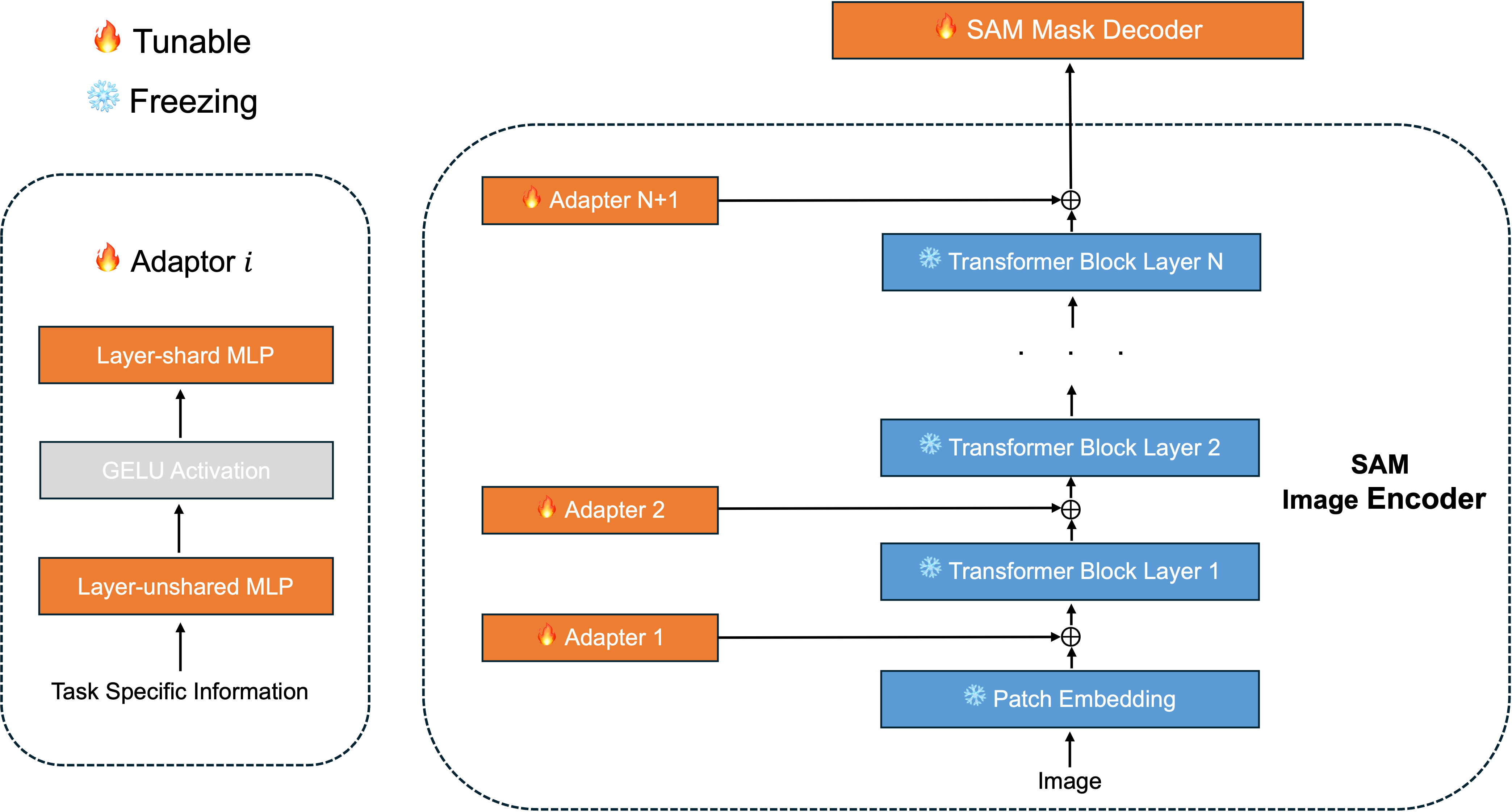}}
    \caption{SAM Adapter architecture.}
    \label{figure_2} 
\end{figure}

In addition to the parallel adapter addition shown in Figure 1, there are also studies that sum the outputs of the transformer block and the adapter. For instance, EVP \cite{EVP} combines patch embedding tuning with the extraction of high-frequency components, which are then input into an lightweight MLP adapter module. The output of this MLP is added to each transformer block. In EVP, the information input to the MLP is termed ``task-specific information". 

The SAM adapter \cite{SA}, which adds adapters to the SAM image encoder as shown in Figure 2, incorporating lightweight MLPs within the ViT along with the use of task-specific information. Besides this, various adapter architectures have been proposed, such as Med-SA \cite{MED-SA} and R-SAM-Seg \cite{RSAMSEG}, which add MLPs after self-attention operations or connect them in parallel to the MLP as shown in Figure 3.

\begin{figure}[!h]
    \centerline{\includegraphics[width=8cm]{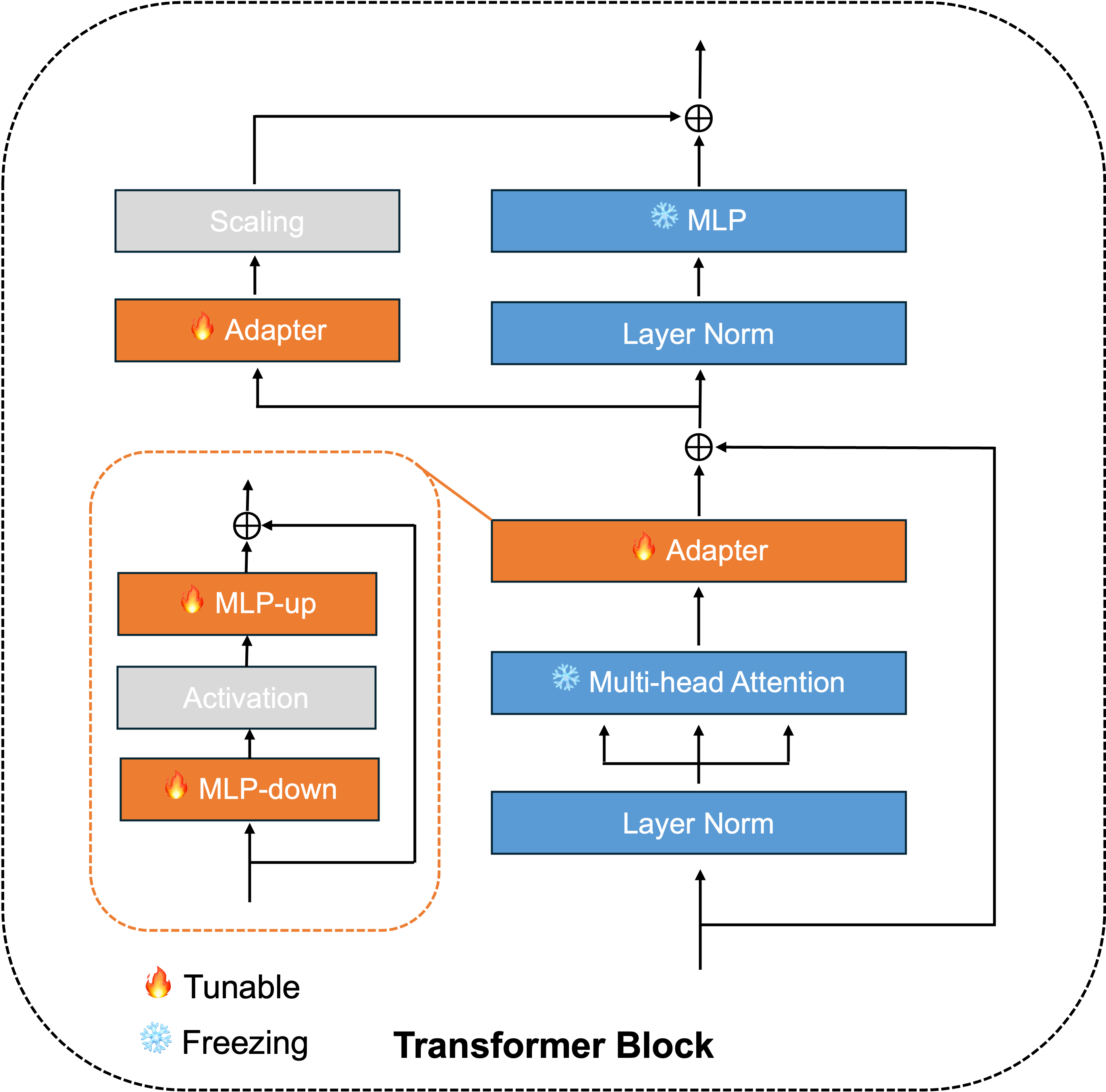}}
    \caption{Adapters in transformer block.}
    \label{figure_3} 
\end{figure}

These adapter module, with their small number of trainable parameters, demonstrate remarkable performance improvements relative to the number of trainable parameters, enabling parameter-efficient fine-tuning. However, due to the adapter's position within the transformer block, training it through backpropagation requires storing the activations and gradients of the transformer block. As evident in Table 1, 2 above, this necessitates an disproportionately large amount of GPU memory relative to the number of parameters, potentially rendering training infeasible on standard GPUs.

\section{Method}

\begin{figure}[h]
    \centerline{\includegraphics[width=15cm]{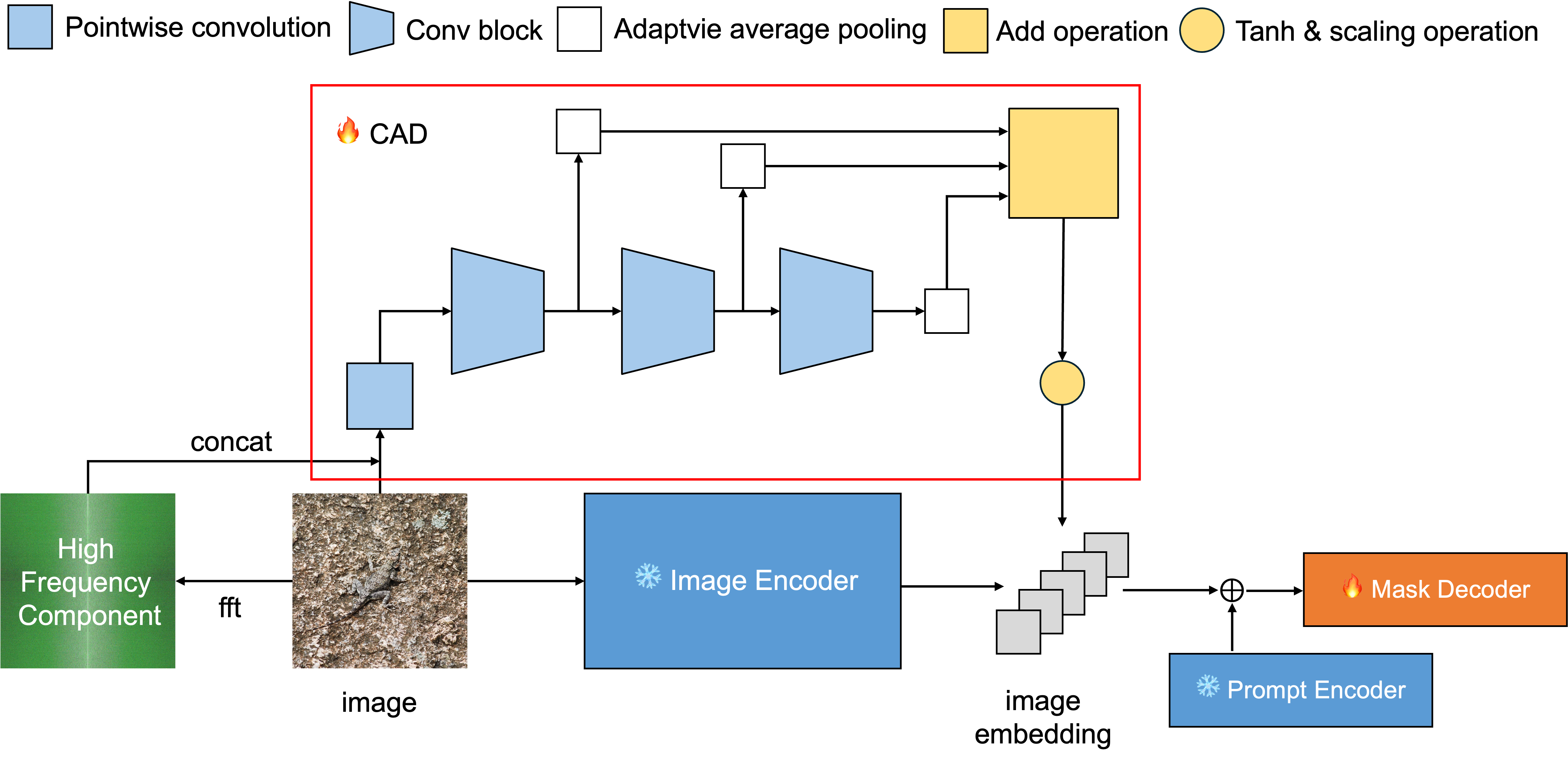}}
    \caption{CAD architecture.}
    \label{figure_4} 
\end{figure}

Figure 4 illustrates our parallel convolutional adapter (CAD) architecture.

\subsection{High Frequency Components (HFC)}

High-frequency components (HFC) are known to be effective in image segmentation, and previous studies \cite{EVP, SA} have utilized HFC for segmentation tasks. Following these precedents, we extract and use the HFC from the input image.

We employ Fast Fourier Transform (FFT) to convert the image into the frequency domain. A zero mask is then applied to the central portion of the transformed image, retaining only the high-frequency components.

\begin{figure}[]
    \centerline{\includegraphics[width=10cm]{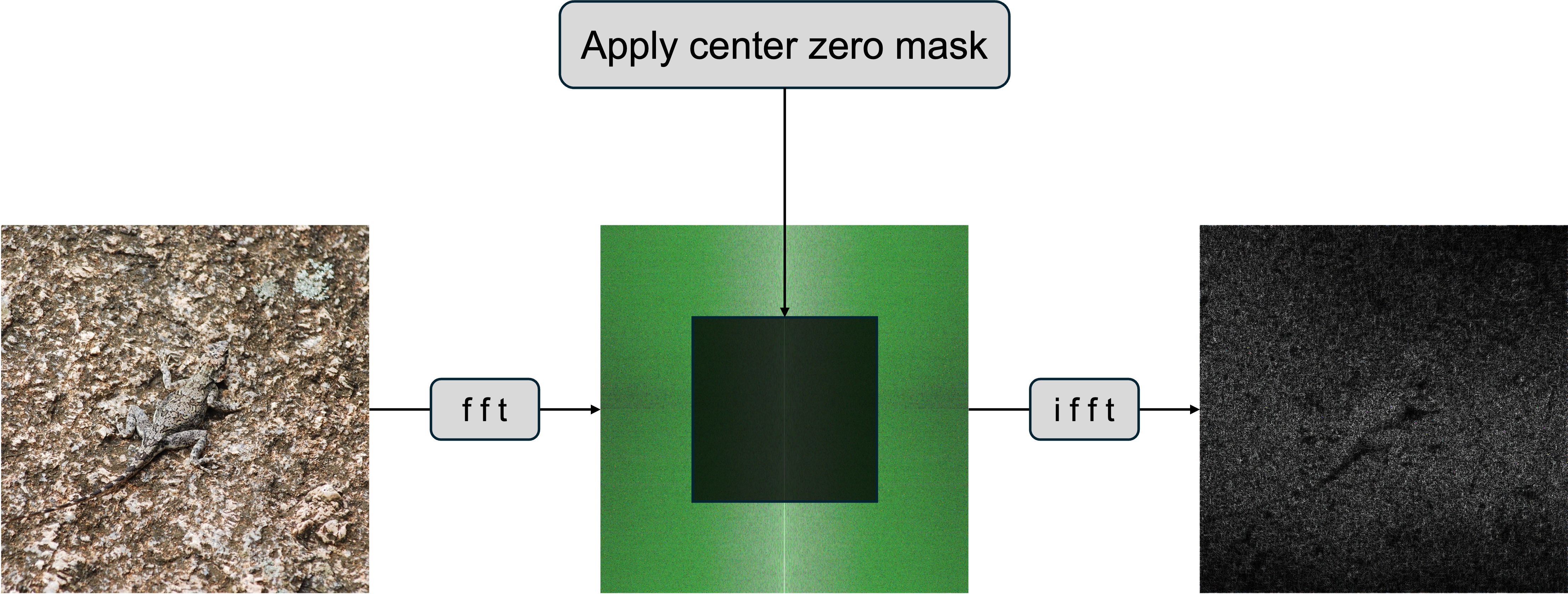}}
    \caption{The process to generate high-frequency components.}
    \label{figure_5} 
\end{figure}

Inverse Fast Fourier Transform (IFFT) is subsequently applied to this high-frequency image to extract the HFC of the input image. These values are then concatenated along the channel dimension of the input image. Consequently, the input to the CAD has dimensions of (batch size, 6, height, width). The value 6 represents the sum of the original image's 3 RGB channels and the 3 RGB channels of the HFC.

\subsection{Convolutional Adapter}

The output of the CAD is designed to have the same dimensions as the image embedding from the image encoder, and this output is subsequently added to the image embedding vector. 

Initially, we normalize each channel by dividing it by its respective maximum value, ensuring all six channels have a distribution within the range [0.0, 1.0]. Subsequently, we adjust the channel dimension using pointwise convolution operations. Then, through three Conv block operations and adaptive average pooling, we adjust the size of the feature map to match that of the image embedding. 

A single Conv block consists of a 3x3 convolution, leaky ReLU activation, batch normalization, and average pooling. This block is designed to halve the size of the feature map with each operation.

\subsection{Output Scaling}

The range of values for image embeddings is approximately [-1.0, 1.0], and the subsequent mask decoder is expected to operate on this input distribution. Considering this, we employed the tanh activation function to constrain the output distribution of the CAD to [-1.0, 1.0]. Additionally, to ensure that the adapter's output does not induce excessive changes to the image embedding, we scaled its magnitude by multiplying it by 0.1.

\section{Experiments}

We trained the model on publicly available shadow detection and camouflaged object detection datasets. Along with our proposed convolutional adapter, we also trained the SAM mask decoder and SAM-adapter to evaluate the performance of each model.

\subsection{Implementation Details}

All models were trained using BCE loss and IoU loss, with the AdamW optimizer. The experimental conditions were set based on those used in the SAM-adapter. For segmentation evaluation metrics, we employed Dice and IoU metrics.

Experiments were conducted using models based on both ViT-B and ViT-H architectures. For ViT-B, a batch size of 4 was employed, while for ViT-H, a batch size of 2 was used. All models were trained for 20 epochs. All the experiments are performed on a single NVIDIA A100 with 80GB memory.

For SAM, we loaded the model using pre-trained checkpoints and then trained only the parameters of specific module such as the adapter and mask decoder. For input prompts, we used box of the same size as the input image. In the case of SAM adapter, we used a model that we implemented ourselves, referencing the official implementation.

\subsection{Datasets}

For shadow detection, we utilized the ISTD dataset, while for camouflaged object detection, we employed the CAMO dataset and the COD10K dataset.

The ISTD dataset comprises 1,330 training images and 540 test images. The CAMO dataset contains 1,000 training images and 250 test images, while the Cod10k dataset includes 3,040 training images and 2,026 test images. As the image sizes varied across these datasets, we resized all images to 512x512 before inputting them into the model.

\subsection{Comparative Results}

Tables 3 and 4 summarize the experimental results, Figures 6 and 7 illustrate the GPU memory usage and Dice scores for each model across the respective datasets. ``SAM" refers to the zero-shot performance without any training on the respective datasets. ``Decoder" indicates results where only the mask decoder of SAM was trained, while ``SA" denotes the SAM Adapter.

\begin{table}[]
\centering
\begin{tabular}{c|cc|cc|cc}
\hline
           & \multicolumn{2}{c|}{ISTD} & \multicolumn{2}{c|}{COD10K} & \multicolumn{2}{c}{CAMO} \\ \cline{2-7} 
           & Dice        & IoU         & Dice         & IoU          & Dice        & IoU        \\ \hline
SAM        & 0.1495      & 0.0889      & 0.0601       & 0.0347       & 0.1637      & 0.1025     \\
Decoder    & 0.7773      & 0.6889      & 0.6764       & 0.5738       & 0.6673      & 0.5484     \\
SA         & 0.9039      & 0.8567      & 0.7343       & 0.6400       & 0.7393      & 0.6281     \\
CAD (Ours) & 0.8681      & 0.8086      & 0.6489       & 0.5409       & 0.6847      & 0.5616     \\ \hline
\end{tabular}
\caption{Quantitative result with ViT-B.}
\label{tab:vit-b}
\end{table}

\begin{table}[]
\centering
\begin{tabular}{c|cc|cc|cc}
\hline
           & \multicolumn{2}{c|}{ISTD} & \multicolumn{2}{c|}{COD10K} & \multicolumn{2}{c}{CAMO} \\ \cline{2-7} 
           & Dice        & IoU         & Dice         & IoU          & Dice        & IoU        \\ \hline
SAM        & 0.2153      & 0.1307      & 0.0498       & 0.0294       & 0.1387      & 0.0879     \\
Decoder    & 0.7544      & 0.6638      & 0.7420       & 0.6579       & 0.7690      & 0.6730     \\
SA         & 0.9516      & 0.9220      & 0.8600       & 0.7890       & 0.8177      & 0.7292     \\
CAD (Ours) & 0.8616      & 0.8069      & 0.7464       & 0.6613       & 0.7578      & 0.6595     \\ \hline
\end{tabular}
\caption{Quantitative result with ViT-H.}
\label{tab:vit-h}
\end{table}

\begin{figure}[]
    \centerline{\includegraphics[width=15cm]{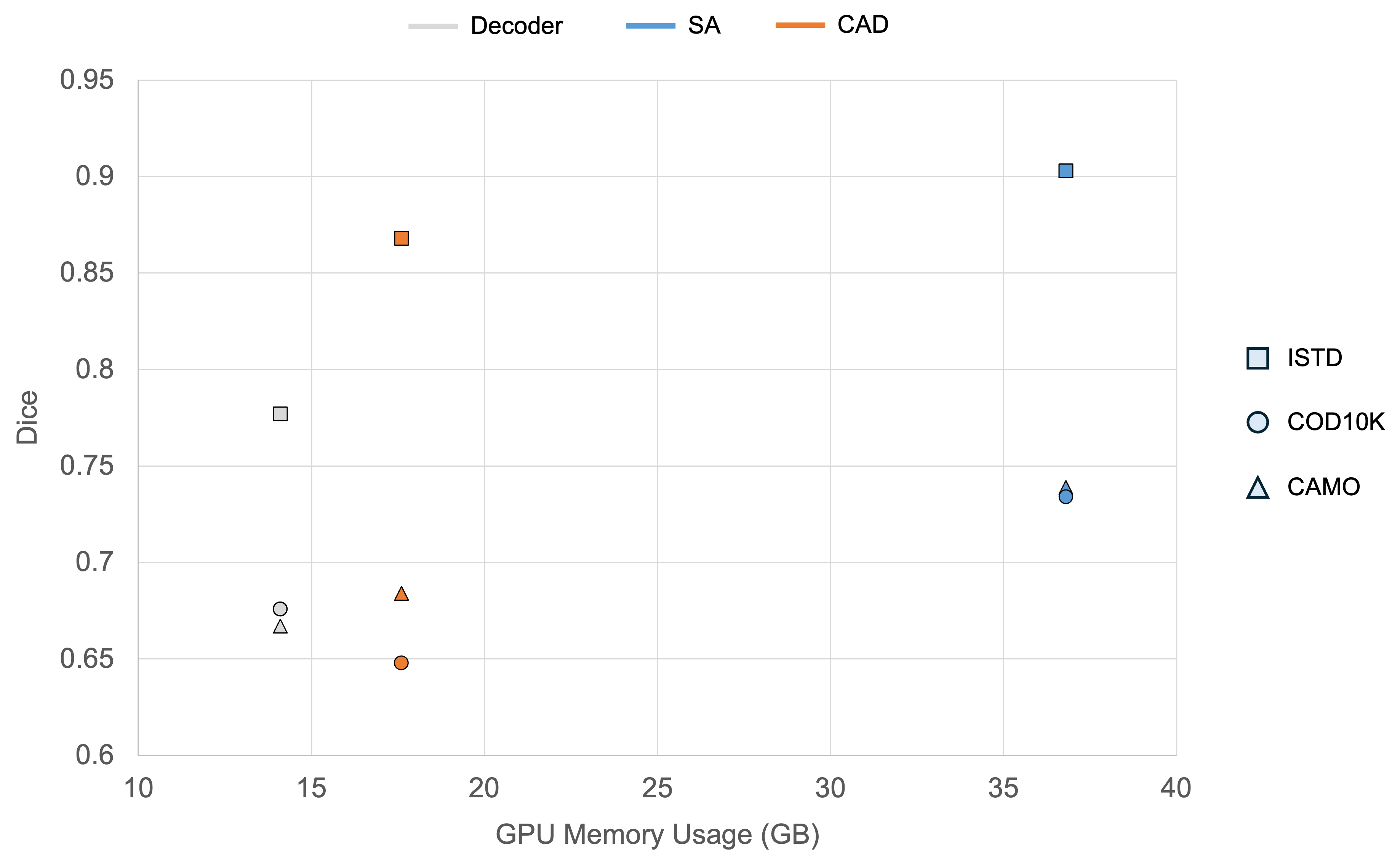}}
    \caption{The dice-memory trade-off for SAM Decoder, SAM Adapter, SAM Conv Adapter with ViT-B.}
    \label{figure_6} 
\end{figure}

\begin{figure}[]
    \centerline{\includegraphics[width=15cm]{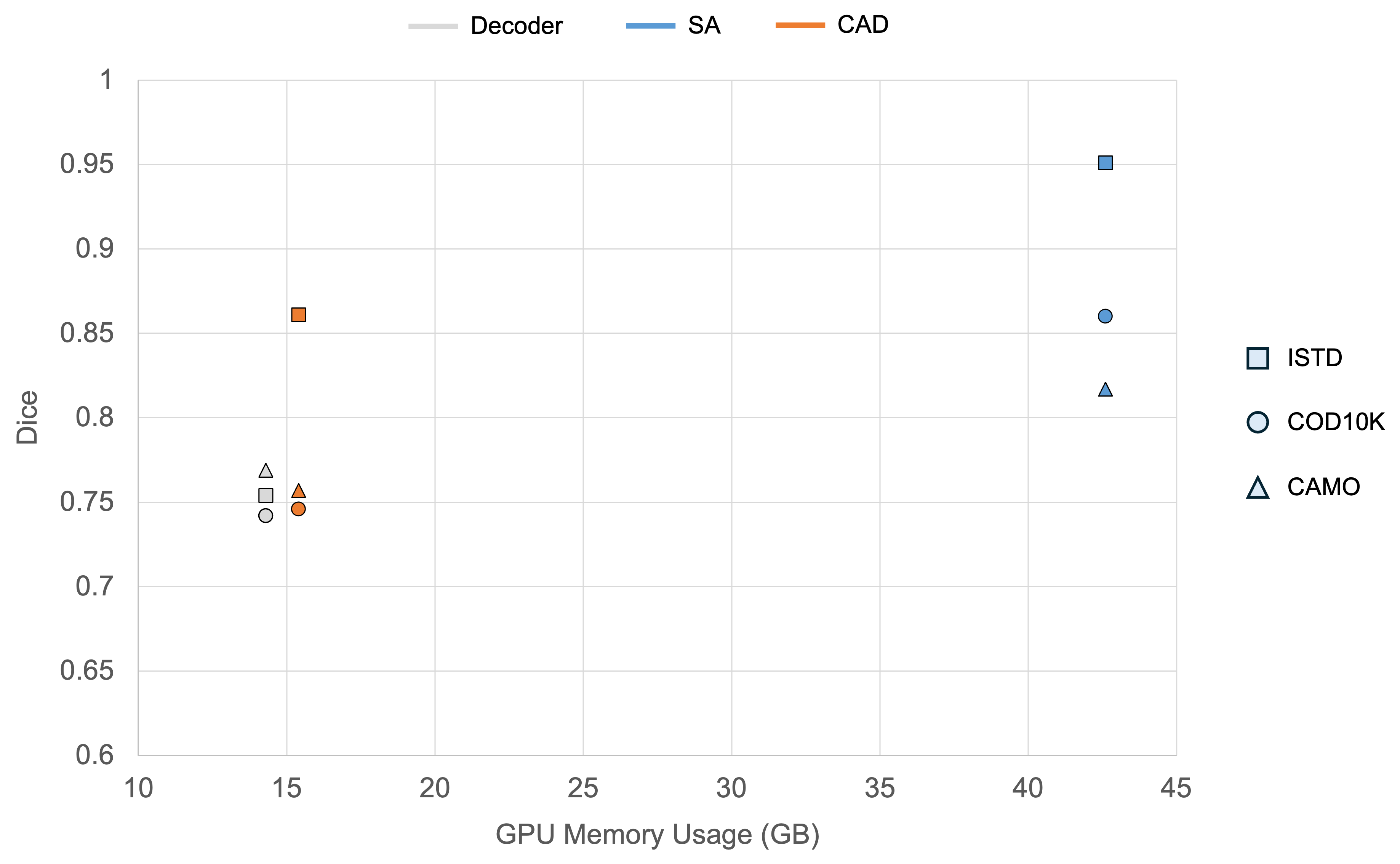}}
    \caption{The dice-memory trade-off for SAM Decoder, SAM Adapter, SAM Conv Adapter with ViT-H.}
    \label{figure_7} 
\end{figure}

Across all evaluated datasets, the zero-shot performance of SAM, without any task-specific training, consistently yielded the lowest performance metrics. This observation held true even when employing the higher-capacity ViT-H image encoder. The significant performance gap between zero-shot and fine-tuned models underscores the inherent limitations of foundation models when applied to specialized tasks without adaptation. This finding emphasizes the critical need for domain-specific fine-tuning to achieve optimal performance in targeted applications. A notable observation was the substantial performance enhancement achieved through the fine-tuning of SAM's mask decoder. This improvement suggests that even minimal task-specific adaptation can yield significant gains in segmentation accuracy.

The SAM Adapter consistently demonstrated superior performance metrics across all model variants and evaluated datasets, suggesting its robust adaptability to diverse segmentation tasks. The transition from ViT-B to ViT-H architecture resulted in notable performance enhancements, indicating a positive correlation between model capacity and segmentation accuracy. In ISTD dataset, the ViT-H-based SAM Adapter achieved a near-perfect Dice score, demonstrating excellent segmentation performance.

Despite its impressive performance, the ViT-H-based model presents significant computational challenges. As illustrated in Figure 7, its training process demands in excess of 40GB of GPU memory. This substantial resource requirement introduces practical limitations: a) Training infeasibility on GPUs with less than 40GB memory, b) Necessity for extremely small batch sizes such as 1 on hardware with limited memory capacity.

\begin{figure}[!h]
    \centerline{\includegraphics[width=14cm]{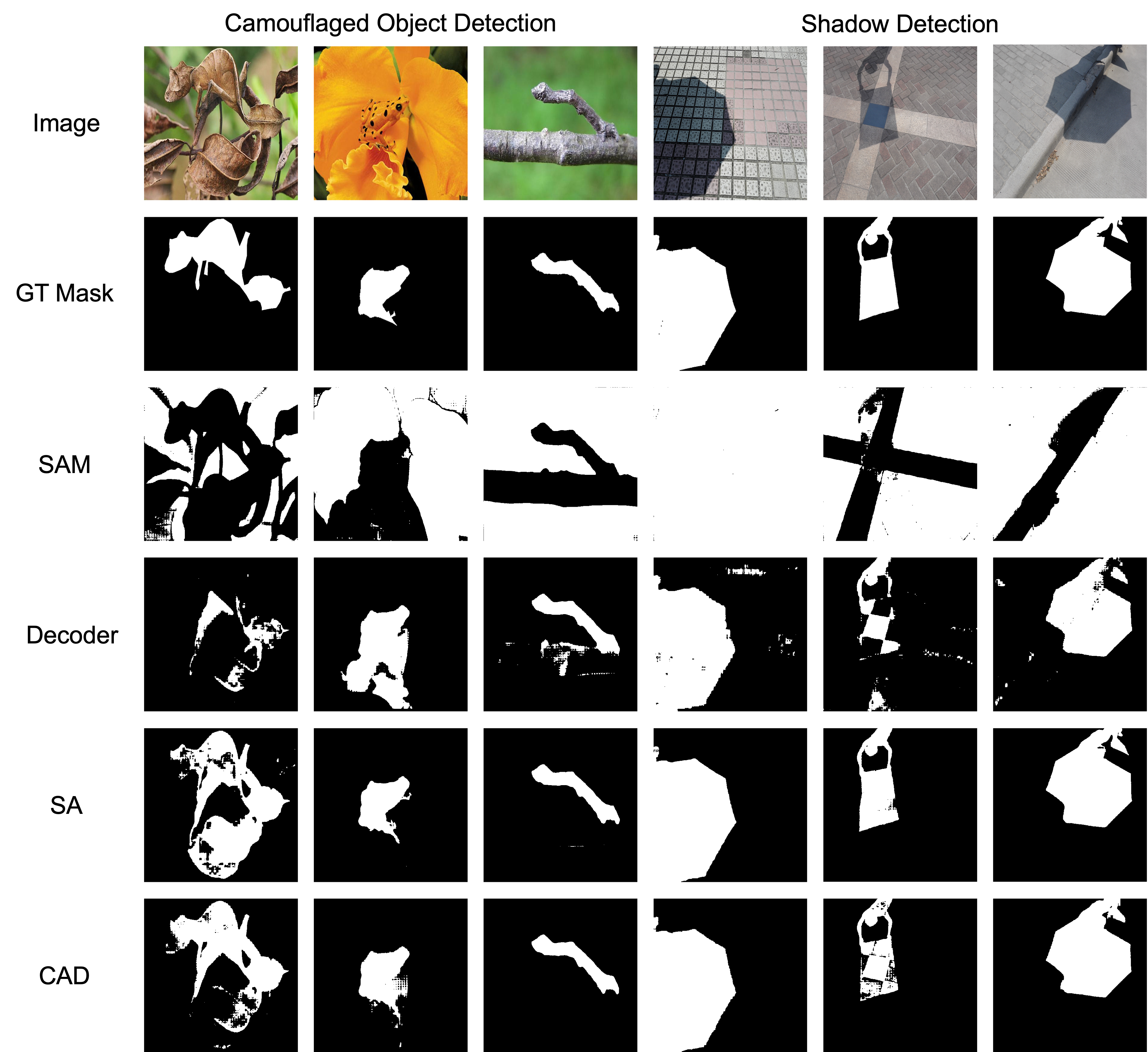}}
    \caption{The visualization results comparing the outputs of each model.}
    \label{figure_8} 
\end{figure}
 
With the exception of two specific cases (COD10K dataset with ViT-B and CAMO dataset with ViT-H), CAD consistently outperformed the baseline approach of simple mask decoder fine-tuning. This trend was particularly pronounced in the ISTD dataset, where a substantial performance increment was observed. As illustrated in Figures 6 and 7, CAD achieved these performance improvements while maintaining a memory footprint comparable to that of mask decoder training. 

The experimental outcomes position CAD as an attractive intermediate solution in the spectrum of adaptation strategies. It offers enhanced performance over simple mask decoder fine-tuning while remaining computationally feasible in scenarios where more resource-intensive approaches like SAM Adapter are impractical due to hardware constraints.

Figure 8 illustrates the output results from each model. It is evident that the vanilla SAM, which was not trained on these specific datasets, either fails to accurately identify the target objects or incorrectly inverts the detection of foreground (object) and background. This observation demonstrates that models, as exemplified by Decoder, SA, and CAD, require dataset-specific training to at least avoid confusing foreground and background elements. While the models may not achieve perfect object identification due to the inherent difficulty of the tasks, it is noteworthy that SA and CAD consistently detect regions that closely resemble the GT Mask in the most cases.

\section{Discussions}

Up to this point, we conducted experiments using the same batch size for both CAD and SAM Adapter to ensure fair experimental conditions. However, given the same GPU memory usage, CAD can actually utilize a batch size nearly twice as large as SAM Adapter. While there isn't a linear relationship between increasing batch size and model performance, using extremely small batch sizes such as 1 or 2 can lead to training instability due to noisy samples, potentially making model convergence impossible.

When using heavier transformer backbones such as ViT-H instead of ViT-B, the number of inserted adapter modules increases. This could potentially lead to a greater performance improvement in SAM Adapter, potentially widening the performance gap between CAD and SAM Adapter. However, employing larger backbones like ViT-H results in a substantial increase in GPU memory usage. Such high memory requirements make training on standard GPUs challenging. Even when training is feasible, it necessitates the use of extremely small batch sizes. Conversely, parallel architecture like the CAD do not significantly increase GPU memory usage even with heavier backbones. The memory usage only increases by the amount of parameters in the enlarged backbone.

Due to the parallel connection of CAD with the image encoder, the learning process only requires the image embedding output by the image encoder, not the image encoder itself. Leveraging this characteristic, one could pre-compute image embeddings for training images and utilize them during the learning process. This approach would eliminate the need to load the image encoder into memory, resulting in significantly reduced GPU memory usage and faster model training in the same environment.

\section{Conclusion}

In this paper, we introduce the GPU memory issues that are not typically discussed in the fine-tuning process using Adapters, and propose a parallel convolutional adapter architecture to address the memory constraints in the training process of Segment Anything. We conducted experiments with the proposed architecture on shadow detection and camouflaged object detection tasks, which are introduced as challenging tasks for Segment Anything. The proposed architecture, CAD, was able to achieve superior performance in most cases compared to simple mask decoder fine-tuning. When more accurate models are required than those obtained through simple mask decoder training, but the addition of adapters is not feasible due to hardware limitations, we anticipate that the CAD proposed in this paper may serve as a viable alternative worth exploring.

\end{document}